\documentclass[sn-mathphys-num]{sn-jnl}

\usepackage{graphicx}%
\usepackage{subcaption}
\usepackage{multirow}%
\usepackage{amsmath,amssymb,amsfonts}%
\usepackage{pifont} 
\usepackage{amsthm}%
\usepackage[scr=boondox]{mathalpha}   
\usepackage[title]{appendix}%
\usepackage{xcolor}%
\usepackage{textcomp}%
\usepackage{manyfoot}%
\usepackage{booktabs}%
\usepackage{algorithm}%
\usepackage{algorithmicx}%
\usepackage{algpseudocode}%
\usepackage{listings}%

\theoremstyle{thmstyleone}%
\theoremstyle{thmstyletwo}%

\theoremstyle{thmstylethree}%
\newtheorem{definition}{Definition}%

\usepackage{stmaryrd} 
\usepackage{dsfont} 
\usepackage{mathabx} 
\usepackage{bbm}

\usepackage{tikz} \usetikzlibrary{calc, arrows.meta, intersections, patterns, positioning, shapes.misc, fadings, through,decorations.pathreplacing}
\tikzstyle{edge} = [fill,opacity=.2,fill opacity=.9,line cap=round, line join=round, line width=50pt]
\usetikzlibrary{fit}

\newcommand{\ie}{\textit{i}.\textit{e}. }
\newcommand{\eg}{\textit{e}.\textit{g}.\ }

\DeclareMathOperator*{\argmax}{argmax}

\usepackage[xcolor]{mdframed}
\definecolor{lightblue}{rgb}{0.96, 0.99, 0.99}
\definecolor{darkblue}{rgb}{0.7,0.81,0.87}
\definecolor{brun}{rgb}{0.87,0.72,0.53}
\definecolor{chamois}{rgb}{1.0,0.90,0.70}
\definecolor{darkpurple}{rgb}{0.81,0.7,0.87}
\definecolor{lightpurple}{rgb}{0.99, 0.96, 0.99}
\definecolor{lightbrown}{rgb}{0.99, 0.99, 0.96}
\definecolor{darkbrown}{rgb}{0.87,0.81,0.7}
\definecolor{dark}{rgb}{0.81,0.87,0.7}

\begin{document}

\title[Neurosymbolic Conformal Classification]{Neurosymbolic Conformal Classification}


\author*[1,2]{\fnm{Arthur} \sur{Ledaguenel}}\email{arthur.ledaguenel@irt-systemx.fr}

\author[2]{\fnm{Céline} \sur{Hudelot}}

\author[1]{\fnm{Mostepha} \sur{Khouadjia}}

\affil[1]{\orgname{IRT SystemX}, \orgaddress{\city{Palaiseau}, \country{France}}}

\affil[2]{\orgdiv{MICS}, \orgname{CentraleSupélec}, \orgaddress{\state{Saclay}, \country{France}}}



\abstract{The last decades have seen a drastic improvement of Machine Learning (ML), mainly driven by Deep Learning (DL). However, despite the resounding successes of ML in many domains, the impossibility to provide guarantees of conformity and the fragility of ML systems (faced with distribution shifts, adversarial attacks, etc.) have prevented the design of trustworthy AI systems. Several research paths have been investigated to mitigate this fragility and provide some guarantees regarding the behavior of ML systems, among which are neurosymbolic AI and conformal prediction. Neurosymbolic artificial intelligence is a growing field of research aiming to combine neural network learning capabilities with the reasoning abilities of symbolic systems. One of the objective of this hybridization can be to provide theoritical guarantees that the output of the system will comply with some prior knowledge. Conformal prediction is a set of techniques that enable to take into account the uncertainty of ML systems by transforming the unique prediction into a set of predictions, called a confidence set. Interestingly, this comes with statistical guarantees regarding the presence of the true label inside the confidence set. Both approaches are distribution-free and model-agnostic. In this paper, we see how these two approaches can complement one another. We introduce several neurosymbolic conformal prediction techniques and explore their different characteristics (size of confidence sets, computational complexity, etc.).}

\keywords{Neurosymbolic, Probabilistic Reasoning, Computational complexity}



\maketitle

\section{Introduction} \label{sec:intro}
The last decades have seen a drastic improvement of Machine Learning (ML), mainly driven by Deep Learning (DL). However, despite the resounding successes of ML in many domains, the impossibility to provide guarantees of conformity and the fragility of ML systems (faced with distribution shifts, adversarial attacks, etc.) have prevented the design of trustworthy AI systems. Several research paths have been investigated to mitigate this fragility and provide some guarantees regarding the behavior of ML systems, among which are neurosymbolic AI and conformal prediction, which are both distribution-free (\ie it does not assume anything about the distribution underlying the data) and model-agnostic (\ie it does not prescribes a particular architecture for the neural network). 

Neurosymbolic Artificial Intelligence (NeSy AI) is a growing field of research aiming to combine neural network learning capabilities with the reasoning abilities of symbolic systems. This hybridization can take many shapes depending on how the neural and symbolic components interact, like shown in \cite{Kautz2022T,wang2023dataand}. An important sub-field of neurosymbolic AI is Informed Machine Learning \cite{VonRueden2023}, which studies how to leverage background knowledge to improve neural systems. There again, proposed techniques in the literature can be of very different nature depending on the type of task (\eg regression, classification, detection, generation, etc.), the language used to represent the background knowledge (\eg mathematical equations, knowledge graphs, logics, etc.), the stage at which knowledge is embedded (\eg data processing, neural architecture design, learning procedure, inference procedure, etc.) and benefits expected from the hybridization (\eg explainability, performance, frugality, etc.). In this paper, we tackle supervised \textbf{classification} tasks informed by prior knowledge represented as a \textbf{propositional formula}. 

Conformal prediction is a set of techniques that enable to take into account the uncertainty of ML systems by transforming the unique prediction into a set of predictions, called a confidence set. Interestingly, this comes with statistical guarantees regarding the presence of the true label inside the confidence set. Computationally prohibitive in its original formulation, called Transductive Conformal Prediction, it was later made more amenable to deep learning systems with Inductive Conformal Prediction, while preserving the statistical guarantee. However, due to the exponential size of the output space, multi-label conformal classification still represents a challenge.

The contributions and outline of the paper are the following. We start with preliminary notions on informed classification and conformal classification in Section \ref{sec:preliminaries}. Then we introduce a new method for multi-label conformal classification in Section \ref{sec:new_conformal}. Finally, we introduce two techniques that integrate prior knowledge in conformal classification in Section \ref{sec:informed_conformal}.

\section{Preliminaries} \label{sec:preliminaries}

\subsection{Informed supervised classification} \label{sec:logics}
In machine learning, the objective is usually to learn a functional relationship $f:\mathcal{X} \mapsto \mathcal{Y}$ between an \textbf{input domain} $\mathcal{X}$ and an \textbf{output domain} $\mathcal{Y}$ from data samples. Supervised multi-label classification is a subset of machine learning where input samples are labeled with subsets of a finite set of classes $\mathbf{Y}$. Therefore, labels can be understood as states on the set of variables $\mathbf{Y}$. In this case, the output space of the task, \ie the set of all labels, is $\mathcal{Y} = \mathbb{B}^{\mathbf{Y}}$. In \textbf{informed} supervised (multi-label) classification, prior knowledge (sometimes called background knowledge) specifies which states in the output domain are semantically \textbf{valid}, \ie to which states can input samples be mapped. The set of valid states constitute a boolean function $\mathscr{f} \in \mathbb{B}^{\mathbb{B}^{\mathbf{Y}}}$ on the set of variables $\mathbf{Y}$. As shown in Section \ref{sec:logics}, a natural way to represent such knowledge is to use a logical theory, \ie to provide an abstract logic $(\mathcal{T}, \mathscr{s})$ and a satisfiable theory $T \in \mathcal{T}(\mathbf{Y})$ such that $\mathscr{s}(T) = \mathscr{f}$. For instance, hierarchical and exclusion constraints are used in \cite{Deng2014}, propositional formulas in conjunctive normal form are used in \cite{Xu2018}, boolean circuits in \cite{Ahmed2022spl}, ASP programs in \cite{Yang2020} and linear programs in \cite{niepert_implicit_2021}.

\subsection{Propositional Logic}
A \textbf{propositional signature} is a set $\mathbf{Y}$ of symbols called \textbf{variables} (\eg $\mathbf{Y} = \{a,b\}$). A \textbf{propositional formula} is formed inductively from variables and other formulas by using unary ($\neg$, which expresses negation) or binary ($\lor, \land$, which express disjunction and conjunction respectively) connectives (\eg $\kappa = a \land b$ which is \textit{true} if both variables $a$ and $b$ are \textit{true}). We note $\mathcal{F}(\mathbf{Y})$ the set of formulas that can be formed in this way. A state $\mathbf{y} \in \mathbb{B}^{\mathbf{Y}}$ can be inductively extended to define a \textbf{valuation} $\mathbf{y}^*$ on all formulas using the standard semantics of propositional logic (\eg $\mathbf{y}^*(a \land b) = \mathbf{y}(a) \times \mathbf{y}(b)$). We say that a state $\mathbf{y}$ \textbf{satisfies} a formula $\kappa$, noted $\mathbf{y} \models \kappa$, if $\mathbf{y}^*(\kappa) = 1$. We say that a formula is \textbf{satisfiable} when it is satisfied by at least one state. We use the symbol $\top$ to represent \textbf{tautologies} (\ie formulas which are satisfied by all states). Two formulas $\kappa$ and $\gamma$ are said \textbf{equivalent}, noted $\kappa \equiv \gamma$, if they are satisfied by exactly the same states. We refer to \cite{Russell2021} for more details on propositional logic.

\subsection{Probabilistic reasoning} \label{sec:probs}
One challenge of neurosymbolic AI is to bridge the gap between the discrete nature of logic and the continuous nature of neural networks. Probabilistic reasoning can provide the interface between these two realms by allowing us to reason about uncertain facts. In this section, we introduce two probabilistic reasoning problems: \textbf{Probabilistic Query Estimation} (PQE), \ie computing the probability of a formula to be satisfied, and \textbf{Most Probable Explanation} (MPE), \ie finding the most probable state that satisfies a given formula.

A probability distribution on a set of \textbf{boolean variables} $\mathbf{Y}$ is an application $\mathcal{P}:\mathbb{B}^{\mathbf{Y}} \mapsto \mathbb{R}^+$ that maps each state $\mathbf{y}$ to a probability $\mathcal{P}(\mathbf{y})$ such that $\sum_{\mathbf{y} \in \mathbb{B}^{\mathbf{Y}}} \mathcal{P}(\mathbf{y})=1$. To define internal operations between distributions, like multiplication, we extend this definition to un-normalized distributions $\mathcal{E}:\mathbb{B}^{\mathbf{Y}} \mapsto \mathbb{R}^+$. The \textbf{null distribution} is the application that maps all states to $0$. The \textbf{partition function} $\mathsf{Z}:\mathcal{E} \mapsto \sum_{\mathbf{y} \in \mathbb{B}^{\mathbf{Y}}} \mathcal{E}(\mathbf{y})$ maps each distribution to its sum, and we note $\overline{\mathcal{E}} := \frac{\mathcal{E}}{\mathsf{Z}(\mathcal{E})}$ the normalized distribution (when $\mathcal{E}$ is non-null). The \textbf{mode} of a distribution $\mathcal{E}$ is its most probable state, ie $\underset{\mathbf{y} \in \mathbb{B}^{\mathbf{Y}}}{\argmax}\mathcal{E}(\mathbf{y})$.

The independent multi-label classification system (see Example \ref{ex:imc}) is build by following the probabilistic interpretation based on the \textbf{exponential probability distribution}, which is parameterized by a vector of logits $\mathbf{a} \in \mathbb{R}^k$, one for each variable in $\mathbf{Y}$, and corresponds to the joint distribution of independent Bernoulli variables $\mathcal{B}(p_i)_{1 \leq i \leq k}$ with $p_i = \mathsf{s}(a_i)$.
\begin{definition}
    Given a vector $\mathbf{a} \in \mathbb{R}^k$, the \textbf{exponential distribution} is:
    \begin{equation}
    \mathcal{E}(\cdot | \mathbf{a}):  \mathbf{y} \mapsto \prod_{1 \leq i\leq k} e^{a_i.y_i}
    \end{equation}
    
    We will note $\mathcal{P}(\cdot | \mathbf{a}) = \overline{\mathcal{E}(\cdot | \mathbf{a})}$ the corresponding normalized probability distribution.
\end{definition}

Typically, when belief about random variables is expressed through a probability distribution and new information is collected in the form of evidence (\ie a partial assignment of the variables), we are interested in two things: computing the probability of such evidence and updating our beliefs using Bayes' rules by conditioning the distribution on the evidence. Probabilistic reasoning allows us to perform the same operations with logical knowledge in place of evidence. Let's assume a probability distribution $\mathcal{P}$ on variables $\mathbf{Y}:= \{Y_j\}_{1\leq j \leq k}$ and a \textbf{satisfiable} propositional formula $\kappa$. Notice that $\mathcal{P}$ defines a probability distribution on the set of states of $\mathbf{Y}$. We also note $\mathbbm{1}_{\kappa}$ the indicator function of $\kappa$ which maps satisfying states to $1$ and others to $0$:
\begin{equation*}
    \mathbbm{1}_{\kappa}(\mathbf{y}) = \left\{
    \begin{array}{ll}
        1 & \mbox{if } \mathbf{y} \models \kappa \\
        0 & \mbox{otherwise}
    \end{array}
\right.
\end{equation*}

\begin{definition}
    The \textbf{probability} of $\kappa$ under $\mathcal{P}$ is:
    \begin{equation}
        \mathcal{P}(\kappa) := \mathsf{Z}(\mathcal{P} \cdot \mathbbm{1}_{\kappa}) = \sum_{\mathbf{y} \in \mathbb{B}^{\mathbf{Y}}} \mathcal{P}(\mathbf{y}) \cdot \mathbbm{1}_{\kappa}(\mathbf{y})
    \end{equation}

    The distribution $\mathcal{P}$ \textbf{conditioned on} $\kappa$, noted $\mathcal{P}(\cdot | \kappa)$, is:
    \begin{equation}
        \mathcal{P}(\cdot | \kappa):= \overline{\mathcal{P} \cdot \mathbbm{1}_{\kappa}}
    \end{equation}
\end{definition}


Since $\mathcal{P}(\cdot | \mathbf{a})$ is strictly positive (for all $\mathbf{a}$), if $\kappa$ is satisfiable then its probability under $\mathcal{P}(\cdot | \mathbf{a})$ is also strictly positive. We note:
    \begin{gather*}
        \mathcal{P}(\kappa | \mathbf{a}):=\mathsf{Z}(\mathcal{P}(\cdot | \mathbf{a}) \cdot \mathbbm{1}_{\kappa}) \\
        \mathcal{P}(\cdot | \mathbf{a}, \kappa):=\frac{\mathcal{P}(\cdot | \mathbf{a}) \cdot \mathbbm{1}_{\kappa}}{\mathcal{P}(\kappa | \mathbf{a})}
    \end{gather*}

Computing $\mathcal{P}(\kappa | \mathbf{a})$ is a \textbf{counting} problem called \texttt{Probabilistic Query Estimation} (\texttt{PQE}). Computing the mode of $\mathcal{P}(\cdot | \mathbf{a}, \kappa)$ is an \textbf{optimization} problem called \texttt{Most Probable Explanation} (\texttt{MPE}). Computing the $k$ most probable states of $\mathcal{P}(\cdot | \mathbf{a}, \kappa)$ is called the top-$k$ problem. Enumerating all states in decreasing order of their probability is the ranked enumeration (\texttt{RankedEnum}) problem. Solving these probabilistic reasoning problems is at the core of many neurosymbolic techniques for informed classification.


Techniques for informed conformal classification that we introduce in this paper are based on two other probabilistic reasoning problems (see Section \ref{sec:informed_conformal}). Computing all satisfying states with a probability $\mathcal{P}(\mathbf{y} | \mathbf{a})$ superior to a given threshold $t$ the \texttt{thresholding} (\texttt{Thresh}) problem. A variant of this is to compute all satisfying states with a probability $\mathcal{P}(\mathbf{y} | \mathbf{a}, \kappa)$ superior to a given threshold $t$, which call \texttt{conditional thresholding} (\texttt{CondThresh}). Notice that solving \texttt{PQE} allows to adapt the threshold of the \texttt{Thresh} problem to solve \texttt{CondThresh}.

\subsection{Conformal classification}
One of the great limitations of Machine Learning algorithms is their lack of guarantee regarding the validity of their predictions. Even when the algorithm is underpinned by a probabilistic interpretation, like often in Deep Learning, many experiments show that these probabilities are poorly calibrated: they do not correspond to the validity of the predictions. Indeed, it is not uncommon that an ML system makes a wrong prediction with a high degree of confidence. This results in a lack of trust in Machine Learning systems and is a major obstacle to their widespread adoption.

Conformal Prediction (CP) is a distribution-free and model agnostic framework that can solve this issue by transforming a Machine Learning algorithm from a point-wise predictor into a conformal predictor that outputs sets of predictions (called confidence sets) guaranteed to include the ground truth with a confidence level $1 - \alpha$, where $\alpha$ is a user-defined miscoverage rate. 

\subsubsection{Transductive Conformal classification}
Assume an input space $\mathcal{X}$, an output space $\mathcal{Y}$ and a collection of samples $(x_i, y_i)_{1 \leq i \leq n}$ i.i.d. according to a probability law $\mathcal{P_{X \times Y}}$. We are given a new input $x_{n+1}$ and we would like to make a confident guess about what could be the corresponding output $Y_{n+1}$ knowing that $(x_{n+1}, Y_{n+1})$ was sampled according to $\mathcal{P_{X \times Y}}$. To do so, we define a non-conformity measure $A : (\mathcal{X} \times \mathcal{Y})^n \times (\mathcal{X} \times \mathcal{Y}) \mapsto \mathbb{R}^+$ such that $A(\{(x_1, y_1), ..., (x_n, y_n)\}, (x_{n+1}, y_{n+1}))$ measures how likely it is that a new sample $(x_{n+1}, y_{n+1})$ is i.i.d. with $\{(x_1, y_1), ..., (x_n, y_n)\}$. For every possible output $y \in \mathcal{Y}$ we compute:
\begin{gather}
    \mu^y_i := A(\{(x_1, y_1), (x_{i-1}, y_{i-1}), (x_{i+1}, y_{i+1}), ..., (x_{n+1}, y)\}, (x_i, y_i)), 1 \leq i \leq n \\
    \mu^y_{n+1} := A(\{(x_1, y_1), (x_{i-1}, y_{i-1}), (x_{i+1}, y_{i+1}), ..., (x_n, y_n)\}, (x_{n+1}, y))
\end{gather}
Then, we compute how likely $(x_{n+1}, y)$ is to belong to the sequence compared to every other sample in the sequence, called the p-value of $y$:
\begin{equation}
    p(y) := p((x_1, y_1), ..., (x_n, y_n), (x_{n+1}, y)) := \frac{|\{1 \leq i \leq n+1 | \mu^y_i \geq \mu^y_{n+1}\}|}{n+1}
\end{equation}
The key property of the p-value is the following:
\begin{equation}\label{eq:tcc_guarantee}
    \mathcal{P_{X \times Y}}(\{y \in \mathcal{Y} | p(y) > \alpha \}) \geq 1 - \alpha
\end{equation}
In other terms, the p-values give us a way to select a set of outputs $C_{\alpha}(x):=\{y \in \mathcal{Y} | p(y) > \alpha \}$ which contains the true label with probability at least $1-\alpha$.

Even though it has no effect on the probabilistic guarantee, the challenge in CP is in the design of the non-conformity measure. If it does not correctly discriminates between conform outputs and non-conform outputs, the confidence sets become too large and do not provide significant information about the true output. For instance, if the non-conformity measure is purely random, the confidence sets will in average be of the size $|\mathcal{Y}| \times (1 - \alpha)$, which corresponds to a uniform prior on the outputs and brings no additional information. On way to design a good conformity-measure is to use a learning algorithm: to compute $A(\{(x_1, y_1), ..., (x_n, y_n)\}, (x_{n+1}, y_{n+1}))$, train a machine learning algorithm on the training set $(x_i, y_i)_{1 \leq i \leq n}$ and use it on $x_{n+1}$ to produce the non-conformity scores (for neural classification systems take $1 - \mathcal{P}(\mathbf{y} | \mathsf{M}_{\theta}(x))$ for instance). The problem when applying this procedure with large datasets and deep learning algorithms is obvious: for each sample to classify, the model must be trained once for each sample in the dataset.

\subsubsection{Inductive conformal classification}
To avoid this, Inductive Conformal Prediction (ICP) split available data into two sets: a training set $\mathcal{D}_{train}=(x_i^t, \mathbf{y}_i^t)_{1 \leq i \leq n_{train}}$ used to train the model and produce the non-conformity measure and a \textbf{calibration} set $\mathcal{D}_{cal}=(x_i^c, \mathbf{y}_i^c)_{1 \leq i \leq n_{cal}}$ used to compare the likeliness of outputs. Therefore, the non-conformity measure $A : (\mathcal{X} \times \mathcal{Y})^n \times (\mathcal{X} \times \mathcal{Y}) \mapsto \mathbb{R}^+$ is replaced with a parametric function $s : \Theta \times \mathcal{X} \times \mathcal{Y})  \mapsto \mathbb{R}^+$ such that $s_{\theta}(x, y)$ represents how likely it is that $(x, y)$ is iid with $\mathcal{D}_{train}$ where the parameters $\theta$ are learned from the training set.

With this method, the model is only trained once then evaluated once for each sample in the calibration and evaluation sets. Besides, non-conformity scores for samples in the calibration set must not be re-computed for every possible output. For a given miscoverage rate $\alpha$, we can compute the $\frac{(n_{cal} + 1)(1- \alpha)}{n_{cal}}$-quantile $q_{\alpha}$ of the non-conformity scores on the calibration set $(s(x_i, y_i, \theta))_{1 \leq i \leq n_{cal}}$. This allows to shortcut the computation of the p-values by directly comparing the value of $\mu^y_{n_{cal}+1}$ to $q_{\alpha}$, which gives us the confidence sets:
\begin{equation}\label{eq:icc_confsets}
    C_{\alpha}(x):=\{y \in \mathcal{Y} | s_{\theta}(x, y) < q_{\alpha} \}
\end{equation}

Fortunately, confidence sets defined defined in Equation \ref{eq:icc_confsets} preserve the statistical guarantee expressed in Equation \ref{eq:tcc_guarantee}, \ie:
\begin{equation} \label{eq:icc_guarantee}
    \mathcal{P}(\mathbf{y}^t \in C(x)) \geq 1 - \alpha
\end{equation}


\subsubsection{Multi-label conformal classification}
Non-conformity measures based on neural networks were first developed for in categorical case \cite{Papadopoulos2002}. Later, \cite{Papadopoulos2014} proposed an adapted version to tackle tasks of multi-label classification using the following non-conformity measure:
\begin{equation*}
    s_{\theta}(x, \mathbf{y}) = \sum_{i=1}^k |y_i - p_i|^d
\end{equation*}
where $p_i=\sigma(\mathsf{M}_{\theta}(x)_i)$ and $d$ controls how sensitive the non-conformity measure is to poorly calibrated scores.

A variant is also proposed to take into account unlikely pairs of classes:
\begin{equation*}
    s_{\theta}(x, \mathbf{y}) = \sum_{i=1}^k |y_i - p_i|^d + \sum_{1 \leq i \leq j \leq k} y_i \cdot y_j \cdot \mu_{i,j}
\end{equation*}
where $p_i=\sigma(\mathsf{M}_{\theta}(x)_i)$ and $\mu_{i,j}$ equals 0 if the classes $Y_i$ and $Y_j$ have been observed together on at least one instance of the training set and 1 otherwise.

However, due the exponential size of the output space $\mathcal{Y} := \mathbb{B}^{\mathbf{Y}}$, testing each possible label to find the confidence set becomes intractable when the number of classes increase.

Therefore, \cite{Wang2015} adopts a Binary Relevance approach by treating each variable as a separate binary classification task. Then, the cross-product of all confidence sets is taken. \cite{Lambrou2016} suggests to relax the exact match requirement and instead look at the minimal distance of the true label to the confidence set using the Hamming Loss.

To deal with this explosion, some paper suggested to modify the conformal guarantee: instead of predicting a confidence set of states (or labelsets) that would contain the ground truth state with a high probability (\eg Equation \ref{eq:conformal}), the conformal predictor would predict a state of which the ground truth state would be a subset with a high probability (\eg Equation \ref{eq:mll_weak_conformal}).

\begin{equation} \label{eq:mll_weak_conformal}
    \mathcal{P}(\mathbf{y}^t \subset C(x)) \geq 1 - \alpha
\end{equation}

A variant proposes to bound the ground truth state between inner and outer states (\eg Equation \ref{eq:mll_bound_conformal}).
\begin{equation} \label{eq:mll_bound_conformal}
    \mathcal{P}( C_{in}(x) \subset \mathbf{y}^t \subset C_{out}(x)) \geq 1 - \alpha
\end{equation}

\section{A new method for conformal multi-label classification} \label{sec:new_conformal}
In this section we introduce several new methods for inductive conformal multi-label classification. We first define a new un-informed method that deals with the exponential size of the output space. Then we introduce two methods that integrate prior knowledge about the classification task.

Our new un-informed method for inductive conformal multi-label classification is based on the following non-conformity measure:
\begin{equation*}
    s_{\theta}(x, \mathbf{y}) = 1 - \mathcal{P}(\mathbf{y} | \mathsf{M}_{\theta}(x))
\end{equation*}
More specifically, we exploit two interesting properties of this non-conformity measure:
\begin{enumerate}
    \item Computing the confidence set $C_{\alpha}(x):=\{y \in \mathcal{Y} | s_{\theta}(x, y) < q_{\alpha} \}$ is equivalent to enumerating all states with a probability superior to a threshold $t_{\alpha} = 1 - q_{\alpha}$. This can be done efficiently in time polynomial in the number of variables and in the number of states.
    \item For a given quantile $q_{\alpha}$ and its corresponding probability threshold $t_{\alpha} = 1 - q_{\alpha}$, the number of states for which probability is superior to $t_{\alpha}$ is bounded by $\frac{1}{t_{\alpha}}$ (because the partition function of the probability distribution $\mathcal{P}(\cdot | \mathsf{M}_{\theta}(x))$ is 1).
\end{enumerate}

In practice, the tractability of such a method is bound by the precision of the underlying model: if the quantile is small, only a few states will have to be checked and the method can be implemented efficiently, but if the quantile is large, then the number of states to check can become prohibitively large.

\section{Informed inductive conformal classification} \label{sec:informed_conformal} 
When a task of multi-label classification is informed by a propositional formula $\kappa$, it is possible to integrate this knowledge into the conformal classification method to tighten the size of confidence sets produced by a given miscoverage rate $\alpha$. We introduce to this end two conformal classification methods that integrate prior knowledge and detail their computational complexity.

\subsection{Semantic filtering}
A first approach works as an extension of any inductive conformal classification method by simply filtering the states that do not satisfy $\kappa$ in the confidence set:
\begin{equation}
    C_{\alpha}^{\kappa}(x):=\{ y \in \mathcal{Y} | s_{\theta}(x, y) < q_{\alpha}, y \models \kappa \}
\end{equation}

Importantly, as long as the consistency hypothesis about the test set (\ie all labels in the test set satisfy the prior knowledge $\kappa$) is maintained, this does not impact the statistical guarantee that comes with conformal prediction.

Since satisfaction of a propositional formula can be done in a time linear in the size of the formula, this approach represents a minor overhead to the un-informed inductive conformal classification method in terms of computational complexity.

Besides, when the chosen non-conformity measure is $s_{\theta}(x, \mathbf{y}) = 1 - \mathcal{P}(\mathbf{y} | \mathsf{M}_{\theta}(x))$, an alternative implementation of filtering consists in computing $C_{\alpha}^{\kappa}(x)$ directly instead of computing $C_{\alpha}(x)$ then apply filtering. This can be done efficiently when $\kappa$ can be compiled into a DNNF circuit of reasonable size and brings down the complexity from $|C_{\alpha}(x)|$ to $|C_{\alpha}^{\kappa}(x)|$

\subsection{Semantic conditioning}
The effectiveness of a given non-conformity measure is greatly determined by its alignment to the model's training. Therefore, when the model is trained using semantic conditioning, the parameters are optimized to minimize the conditioned negative likelyhood, and the standard negative likelihood often becomes uninformative. Therefore, prior knowledge can be integrated in the non-conformity measure itself (by conditioning the probability distribution on the prior knowledge) to restore the alignment, \ie:
\begin{equation*}
    s_{\theta}^{\kappa}(x, \mathbf{y}) = 1 - \mathcal{P}(\mathbf{y} | \mathsf{M}_{\theta}(x), \kappa)
\end{equation*}

Because this new non-conformity measure is also based on a probability distribution, it retains the upper bound on the size of the confidence set for a given probability threshold. However, on the contrary to filtering, computing the non-conformity measure and the confidence set is no longer tractable in general. For propositional formulas that can be compiled in dDNNF of reasonable sizes however, such computation can be done tractably in size polynomial in the size of the compiled dDNNF and in the size of the confidence set.


\bibliography{main}

\end{document}